\documentclass[12pt]{llncs}

\usepackage{xspace}
\usepackage{latexsym}
\usepackage{amsmath,amssymb,amscd}

\newcommand{\skippar}[1]{}

\newcommand{\ie}{i.~e.~}

\newcommand{\no}{\ensuremath{\sim}}

\newcommand{\dlp}{\textsl{dlp}}

\newcommand{\strongleft}{\ensuremath{<\hspace{-2mm}\text{---}\hspace{-1mm}\text{---}}}
\newcommand{\strongleftarrow}{{\raise1.5pt\hbox{\tiny\strongleft}}}
\newcommand{\srule}[2]{\ensuremath{#1\;\gets\;#2}}

\newcommand{\facto}[1]{\ensuremath{#1}}
\newcommand{\defleft}{\text{---\hspace{-1.5pt}\raise.05pt\hbox{$<$}}\xspace}
\newcommand{\defleftarrow}{{\raise1.5pt\hbox{\tiny\defleft}}}
\newcommand{\drule}[2]{\ensuremath{#1\; \defleftarrow \; #2}}

\newcommand{\presum}[1]{\drule{#1}{true}}

\newcommand{\defrightarrow}{\ensuremath{\succ\!\text{---}}\xspace}
\newcommand{\defarrow}{{\raise1.5pt\hbox{\tiny\defrightarrow}}}

\newcommand{\SSet}{\ensuremath{\Pi}\xspace}

\newcommand{\DD}{\ensuremath{\Delta}\xspace}
\newcommand{\SD}{\ensuremath{(\SSet,\DD)}\xspace}

\newcommand{\Infer}{\ensuremath{\vdash}\xspace}

\newcommand{\Tree}[1]{\ensuremath{{\mathcal T}_{\tiny #1}}\xspace}

\newcommand{\Argum}[2]{\ensuremath{\langle #1, #2 \rangle }\xspace}

\newcommand{\Co}[1]{\ensuremath{Co(#1)}\xspace}

\newcommand{\Prolog}{P\textsc{rolog}}

\newcommand{\DELP}{\textsl{DeLP}\xspace}

\def\nat{{\ifmmode{\rm I}\mkern-3.5mu{\rm N}
    \else\leavevmode\hbox{I}\kern-.16em \hbox{N}\fi} }

\newcommand{\progP}{\ensuremath{\mathcal{P}}\xspace}

\newcommand{\LAMBDA}{\mbox{\large$\lambda$}}
\newcommand{\linarg}{\mbox{\large$\lambda$}}

\newenvironment{Algo}{\begin{Algox}{\em } \em }{\end{Algox}}
\newtheorem{Algox}{{\sc Algorithm}}[section]

\newcommand{\cpl}[1]{{\overline{#1}}}

\begin{document}

\title{Pruning Search Space in Defeasible Argumentation\protect\footnote{Published
       in Proc. of the Workshop on Advances and Trends in Search
         in Artificial Intelligence, pp.40-47. International Conf. of the Chilean Society
          in Computer Science, Santiago, Chile, 2000.}}

\author{Carlos Iv\'an Ches{\~n}evar \ \ \   Guillermo Ricardo Simari \ \ \ Alejandro Javier Garc\'{\i}a}

\institute{ Universidad Nacional del Sur, Av. Alem 1253, (8000)
  Bahía Blanca, ARGENTINA \\
  Email: {\tt \{cic,grs,ajg\}@cs.uns.edu.ar} }


\maketitle

\begin{abstract}
Defeasible argumentation has experienced a considerable growth in AI
in the last decade. Theoretical results have been combined with
development of practical applications in AI \& Law, Case-Based
Reasoning and various knowledge-based systems.
However,  the dialectical process associated with inference
is computationally expensive.
This paper focuses on speeding up this inference process
by pruning the involved search space. Our approach is twofold. On
one hand, we identify distinguished literals for computing defeat.
On the other hand, we restrict ourselves to a subset of all possible
conflicting arguments by introducing dialectical constraints.
\end{abstract}
\section{Preliminaries}

Argumentation systems (AS) have emerged during the last decade
as a promising formalization of defeasible
reasoning~\cite{Simari92,PraVre99,KakasToni99}.
Starting from the non-monotonic reasoning community, AS evolved
and matured within several areas of Computer Science
such as AI \& Law, knowledge representation, default reasoning and
logic programming.

The inference process in AS is computationally expensive when compared with
alternative frameworks for modeling commonsense reasoning, such as
traditional rule-based systems.
This paper discusses theoretical considerations that lead
to obtain efficient implementations
of AS. As a basis for our analysis we use
\emph{defeasible logic programming}~\cite{GarciaMSC97,GarciaSimariChesnevar98}.
The paper is structured as follows: section~\ref{Sec:DELP}
introduces \emph{defeasible logic programming}.
Section~\ref{sec:pruning}
presents the main contributions of the paper.
Finally, section~\ref{sec:conclusions} concludes.

\section{Defeasible Logic Programming: fundamentals}
\label{Sec:DELP}

Defeasible Logic Programming (\DELP) is a logic programming formalism
which relies upon defeasible argumentation for solving queries.
The \DELP language \cite{Simari92,GarciaMSC97,GarciaSimariChesnevar98}
is defined in terms of two disjoint sets of rules:
\emph{strict rules} for representing strict (sound) knowledge, and
 \emph{defeasible rules} for representing tentative information.
Rules will be defined using \emph{literals}. A literal $L$ is an atom
$p$ or a negated atom ${\no}p$, where the symbol ``$\no$'' represents
\emph{strong negation}.

\begin{definition}[Strict and Defeasible Rules]
  A \emph{strict rule} (\emph{defeasible rule}) is an ordered pair,
  conveniently denoted by $\srule{Head}{Body}$ ($\drule{Head}{Body}$),
  where $Head$, is a literal, and
  $Body$ is a finite set of literals. A strict rule (defeasible rule)
  with the head $L_0$ and body $\{L_1,\ldots,L_n\}$ can also be written as
  $\srule{L_0}{L_1,\ldots,L_n}$ ($\drule{L_0}{L_1,\ldots,L_n}$). If the body
  is empty, it is written $\srule{L}{true}$ ($\presum{L}$), and it is
  called a \emph{fact} (\emph{presumption}). Facts may also be written
  as $\facto{L}$.
\end{definition}

\begin{definition}[Defeasible Logic Program \progP]
\label{program}
A \emph{defeasible logic program} (\dlp) is a finite set of strict and
defeasible rules. If \progP is a \dlp, we will distinguish in \progP
the subset \SSet of strict rules, and the subset \DD of defeasible
rules.  When required, we will denote \progP as \SD.
\end{definition}


\begin{example}
\label{EjemEngineOne}
Consider an agent which has to control
an engine whose performance is determined by three
switches $sw1$, $sw2$ and $sw3$.\footnote{For the sake
of simplicity, we restrict ourselves to propositional language for this example.}
The switches regulate different features of the engine's
behavior, such as pumping system and working speed.
We can model the engine behavior using a \dlp\ program $(\Pi,\Delta)$,
where $\Pi$ = $\{ (\srule{sw1}{}),
                  (\srule{sw2}{}),
                  (\srule{sw3}{}),
                  (\srule{heat}{}),
                  (\srule{{\no}fuel\_ok }{pump\_clogged})\}$
(specifying that the three switches are on, there is heat,
 and whenever the pump gets clogged, fuel is not ok),
and $\Delta$ models the possible behavior of the engine under different
conditions (fig.~\ref{FigDelta}).




\begin{figure}[t]
\begin{tabular}{ll}
\drule{pump\_fuel\_ok}{sw1} & (when sw1 is on, normally fuel is pumped
properly); \\
\drule{fuel\_ok}{pump\_fuel\_ok } &
(when fuel is pumped, normally fuel works ok); \\
\drule{pump\_oil\_ok}{sw2} &
(when sw2 is on, normally oil is pumped); \\
\drule{oil\_ok}{pump\_oil\_ok} &
(when oil is pumped, normally oil works ok); \\
\drule{engine\_ok}{fuel\_ok,oil\_ok } &
(when there is fuel and oil, normally engine works ok); \\
\drule{{\no}engine\_ok}{fuel\_ok,oil\_ok,heat} &
(when there is fuel, oil and heat,
usually engine is not working ok); \\
\drule{{\no}oil\_ok}{heat} & (when there is heat, normally oil is not ok); \\
\drule{pump\_clogged}{pump\_fuel\_ok, low\_speed} &
\begin{tabular}[t]{l}
(when fuel is pumped and speed is low, there are  \\
reasons to believe that the pump is clogged); \\
\end{tabular} \\
\drule{low\_speed}{sw2} & (when sw2 is on, normally speed is low); \\
\drule{{\no}low\_speed}{sw2,sw3} & (when both sw2 and sw3 are
on, speed tends not to be low). \\
\drule{fuel\_ok}{sw3} & (when sw3 is on, normally fuel is ok). \\
\end{tabular}
\caption{Set $\Delta$ (example~\protect\ref{EjemEngineOne})}
\label{FigDelta}
\end{figure}
\end{example}

Given a \dlp\ \progP, a {\em defeasible derivation} for a query
$q$ is a finite set of rules  obtained by backward
chaining from $q$ (as in a \Prolog\ program) using both strict and
defeasible rules from \progP. The symbol ``\no'' is
considered as part of the predicate when generating a defeasible
derivation.
A set of rules $\cal S$ is {\em contradictory} iff there is a defeasible
derivation from $\cal S$ for some literal $p$ and its complement ${\no}p$.
Given a \dlp\ \progP,
we will assume that its set $\Pi$ of strict rules is
non-contradictory.\footnote{If a contradictory set of strict rules is
used in a \dlp\ the same problems as in extended logic programming would appear.
The corresponding analysis has been done elsewhere~\cite{Gelfond90}.}

\begin{definition}[Defeasible Derivation Tree]
  Let \progP be a \dlp, and let $h$ be a ground literal.  A
  {\em defeasible derivation tree} $T$ for $h$ is a finite tree, where
all nodes are labelled with literals, satisfying the following
conditions:
\begin{enumerate}
\item The root node of $T$ is labelled with $h$.
\item For each node $N$ in $T$ labelled with the literal $L$, there
  exists a ground instance of a strict or defeasible rule $r \in
  \progP$ with head $L_0$ and body $\{ L_1, L_2, \ldots, L_k \}$ in $\cal
  P$, such that $L=L\sigma$ for some ground variable substitution $\sigma$,
  and the node $N$ has exactly $k$ children nodes labelled as $L_1\sigma,
  L_2\sigma, \ldots, L_k\sigma$.
\end{enumerate}
The sequence $S$=$[r_1, r_2, \ldots r_k]$ of grounded instances of strict
and defeasible rules used in building $T$ will be called a defeasible
derivation of $h$.
\end{definition}

\begin{definition}[Argument/Subargument]
  Given a \dlp\ \progP, an {\em argument} $\mathcal{A}$ for a query $q$,
denoted \Argum{\mathcal{A}}{q}, is a subset of ground instances of the
defeasible rules of \progP, such that:
  1) there exists a defeasible derivation for $q$ from
        $\Pi \cup  \mathcal{A}$ (also written $\Pi \cup {\cal A} \vdash q$);
  2) $\Pi\cup \mathcal{A}$ is non-contradictory, and
  3) $\mathcal{A}$ is minimal with respect to set inclusion.
An argument \Argum{\mathcal{A}_1}{q_1} is a {\em sub-argument} of another
argument \Argum{\mathcal{A}_2}{q_2}, if $\mathcal{A}_1 \subseteq \mathcal{A}_2$.
\end{definition}


\begin{definition}[Counterargument / Attack]
\label{DefCounterargument}
An argument \Argum{\mathcal{A}_1}{q_1} {\em counterargues} (or \emph{attacks})
an argument \Argum{\mathcal{A}_2}{q_2} at a literal $q$ iff there is an subargument
\Argum{\mathcal{A}}{q} of \Argum{\mathcal{A}_2}{q_2} such that the set $\Pi
\cup \{ q_1,q\}$ is contradictory.
\end{definition}

Informally, a query $q$ will succeed if the supporting argument is not
defeated; that argument becomes a {\em justification}. In order to establish if
$\mathcal{A}$ is a non-defeated argument, \emph{defeaters}
 for $\mathcal{A}$ are considered, \ie counterarguments
that are preferred to $\mathcal{A}$ according to some preference criterion. \DELP
considers a particular criterion called
{\em specificity}~\cite{Simari92,GarciaSimariChesnevar98} which favors an
argument with greater information content and/or less use of
defeasible rules.\footnote{See~\cite{GarciaSimariChesnevar98} for
  details.}

\begin{definition}[Proper Defeater / Blocking Defeater]
\label{DefDefeater}
  An argument \Argum{\mathcal{A}_1}{q_1} defeats \Argum{\mathcal{A}_2}{q_2}
  at a literal $q$ iff there exists a subargument \Argum{\mathcal{A}}{q}
  of \Argum{\mathcal{A}_2}{q_2} such that \Argum{\mathcal{A}_1}{q_1}
  counterargues \Argum{\mathcal{A}_2}{q_2} at $q$, and either: (a)
  \Argum{\mathcal{A}_1}{q_1} is ``better'' that \Argum{\mathcal{A}}{q} (then
  \Argum{\mathcal{A}_1}{q_1} is a proper defeater of \Argum{{\cal
      A}}{q}); or (b) \Argum{\mathcal{A}_1}{q_1} is unrelated by the
  preference order to \Argum{\mathcal{A}}{q} (then \Argum{{\cal
      A}_1}{q_1} is a blocking defeater of \Argum{\mathcal{A}}{q}).
\end{definition}

Since defeaters are arguments, there may exist defeaters for the
defeaters and so on. That prompts for a complete dialectical analysis
to determine which arguments are ultimately defeated. Ultimately
undefeated arguments will be labelled as \emph{U-nodes}, and the
defeated ones as \emph{D-nodes}.  Next we state the formal
definitions required for this process:

\begin{definition}[Dialectical Tree. Argumentation line]
  Let $\mathcal{A}$ be an argument for $q$. A dialectical tree for
  \Argum{\mathcal{A}}{q}, denoted \Tree{\Argum{\mathcal{A}}{q}}, is
  recursively defined as follows:
\begin{enumerate}
\item A single node labeled with an argument \Argum{\mathcal{A}}{q} with
  no defeaters is by itself the dialectical tree
  for \Argum{\mathcal{A}}{q}.
\item Let $\Argum{\mathcal{A}_1}{q_1}, \Argum{\mathcal{A}_2}{q_2}, \ldots,
  \Argum{\mathcal{A}_n}{q_n}$ be all the defeaters (proper or blocking)
  for \Argum{\mathcal{A}}{q}. We construct the dialectical tree for
  \Argum{\mathcal{A}}{q}, \Tree{\Argum{\mathcal{A}}{q}}, by labeling the
  root node with \Argum{\mathcal{A}}{q} and by making this node the
  parent node of the roots of the dialectical trees for $\Argum{{\cal
      A}_1}{q_1}, \Argum{\mathcal{A}_2}{q_2}, \ldots, \Argum{{\cal
      A}_n}{q_n}$.
\end{enumerate}
A path $\lambda$ = [ \Argum{\mathcal{A}_0}{q_0},\ldots
                     \Argum{\mathcal{A}_m}{q_m} ] in
\Tree{\Argum{\mathcal{A}}{q}} is called \emph{argumentation line}.
We will denote as $S_{\lambda}$
= $\bigcup_{i=2k} \Argum{\mathcal{A}_i}{q_i}$
($I_{\lambda}$ = $\bigcup_{i=2k+1} \Argum{\mathcal{A}_i}{q_i}$)
the set of all even-level (odd-level) arguments in $\lambda$.
Even-level (odd-level) arguments are also called
\emph{supporting arguments} or $S$-arguments (\emph{interferring} arguments or
$I$-arguments).
\end{definition}

\begin{definition}[Labelling of the Dialectical Tree]
  Let \Argum{\mathcal{A}}{q} be an argument and \Tree{\Argum{{\cal
        A}}{q}} its dialectical tree, then:
\begin{enumerate}
\item All the leaves in \Tree{\Argum{\mathcal{A}}{q}} are labelled as
  $U$-nodes.
\item Let \Argum{{\cal B}}{h} be an inner node of \Tree{\Argum{{\cal
        A}}{q}}. Then \Argum{{\cal B}}{h} will be a $U$-node iff every
  child of \Argum{{\cal B}}{h} is a $D$-node. The node \Argum{{\cal
      B}}{h} will be a $D$-node iff it has at least one child marked as
  $U$-node.
\end{enumerate}
\end{definition}
To avoid {\em fallacious
argumentation\/}~\cite{Chile94}, two additional constraints on
dialectical trees are imposed on any argumentation line $\lambda$:
a) there can be no repeated arguments
(circular argumentation) and b)
the set of all odd-level (even-level) arguments in $\lambda$
should be \emph{non-contradictory} wrt $\Pi$ in order
to avoid {\em contradictory}  argumentation.
Defeaters satisfying these constraints are
called \emph{acceptable}.\footnote{See~\cite{GarciaSimariChesnevar98} for
an in-depth analysis.}
An argument $\mathcal{A}$ which turns to be ultimately
undefeated is called a {\em justification}. 
\begin{definition}[Justification]
  Let $\mathcal{A}$ be an argument for a literal $q$, and let
  $\Tree{\Argum{\mathcal{A}}{q}}$ be its associated acceptable
  dialectical tree.  The argument $\mathcal{A}$ for $q$ will be a
  {\em justification} iff the root of $\Tree{\Argum{\mathcal{A}}{q}}$ is
a $U$-node.
\end{definition}

\begin{example}
\label{EjemEngineTwo}
Consider example~\ref{EjemEngineOne}, and assume
our agent is trying to determine whether
the engine works ok
by finding a justification supporting $engine\_ok$.
The set of defeasible rules
      $\cal A$ = \{ \drule{pump\_fuel\_ok}{sw1},
                    \drule{pump\_oil\_ok}{sw2},
                    \drule{fuel\_ok}{pump\_fuel\_ok},
                    \drule{oil\_ok}{pump\_oil\_ok},
                    \drule{engine\_ok}{fuel\_ok, oil\_ok} \}.
is an {\em argument} for $engine\_ok$, \ie,
\Argum{{\cal A}}{engine\_ok}.
But there exists a
{\em counterargument}
  $\cal B$ = \{   \drule{pump\_fuel\_ok}{sw1},
                  \drule{low\_speed}{sw2},
                  \drule{pump\_clogged}{pump\_fuel\_ok, low\_speed}  \}
which supports the conclusion ${\no}fuel\_ok$
$(\Pi\ \cup\ \mathcal{B} \Infer\ {\no}fuel\_ok)$.
The argument \Argum{\mathcal{B}}{{\no}fuel\_ok} {\em defeats} \Argum{{\cal A}}{engine\_ok},
since it is more specific.
Hence, the argument
\Argum{{\cal A}}{engine\_ok} will be provisionally rejected, since it is defeated.
However, \Argum{{\cal A}}{engine\_ok}\ can
be {\em reinstated}, since there exists a third argument
${\cal C}=\{ \drule{{\no}low\_speed}{sw2, sw3}\}$
for ${\no}low\_speed$  which on its turn
defeats \Argum{{\cal B}}{{\no}fuel\_ok}.
Note that the argument \Argum{{\cal D}}{fuel\_ok}
with ${\cal D}=\{ \drule{fuel\_ok}{sw3}\}$ would be
also a (blocking) defeater for \Argum{\mathcal{B}}{{\no}fuel\_ok}.

Hence, \Argum{\mathcal{A}}{engine\_ok}\ comes to be undefeated,
since the argument \Argum{\mathcal{B}}{{\no}fuel\_ok} was defeated.
But there is another defeater for \Argum{\mathcal{A}}{engine\_ok},
the argument \Argum{\mathcal{E}}{{\no}engine\_ok}, where
$\cal E$ = \{ \drule{pump\_fuel\_ok}{sw1},
              \drule{pump\_oil\_ok}{sw2},
              \drule{fuel\_ok}{pump\_fuel\_ok},
              \drule{oil\_ok}{pump\_oil\_ok},
              \drule{{\no}engine\_ok}{fuel\_ok, oil\_ok, heat}\}.
Hence \Argum{\mathcal{A}}{engine\_ok} is once again provisionally defeated.

The agent might try to find a defeater for \Argum{\mathcal{E}}{{\no}engine\_ok}\
which could help {\em reinstate} the original argument \Argum{\mathcal{A}}{ok},
for example \Argum{\{ \drule{{\no}oil\_ok}{heat}\}}{{\no}oil\_ok}.
It must be noted, however, that this last argument would be
{\em fallacious}, since there would exist
odd-level supporting arguments for both $oil\_ok$
(as a subargument of \Argum{{\cal A}}{engine\_ok})
and for ${\no}oil\_ok$ (in \Argum{\{ \drule{{\no}oil\_ok}{heat}\}}{{\no}oil\_ok}).
Hence \Argum{\{\drule{{\no}oil\_ok}{heat}\}}{{\no}oil\_ok}\
should not be accepted as a valid defeater for \Argum{\mathcal{A}}{engine\_ok}.
Since there are no more arguments to consider, \Argum{\mathcal{A}}{engine\_ok}\
turns out to be ultimately defeated, so that we can conclude that
the argument \Argum{\mathcal{A}}{engine\_ok}\ is not {\em justified}.
Thus, we conclude that the engine is not working ok.
The argument \Argum{\mathcal{E}}{{\no}engine\_ok}, on its turn,
is a {\em justification}.

Fig.~\ref{FigDialecticalTrees}(b)-left shows the resulting
dialectical tree. Note that \Argum{\mathcal{A}}{engine\_ok}\ is a level-0
supporting argument, and both \Argum{\mathcal{C}}{{\no}low\_speed}\
and \Argum{\mathcal{D}}{fuel\_ok}\ are level-2 supporting arguments.
Both  \Argum{\mathcal{B}}{{\no}fuel\_ok}\ and
\Argum{\mathcal{E}}{{\no}engine\_ok}\
are level-1 interfering arguments.
$\lambda$ = [ \Argum{\mathcal{A}}{engine\_ok},
\Argum{\mathcal{B}}{{\no}fuel\_ok},
\Argum{\mathcal{C}}{{\no}low\_speed}]
is an {\em argumentation line}.
\end{example}

\newcommand{\alfabeta}{\mbox{$\alpha$-$\beta$}}
\section{Pruning dialectical trees}
\label{sec:pruning}

Building a dialectical tree is computationally expensive:
arguments are proof trees, and a dialectical tree is
a tree of arguments. In both cases, consistency checks are needed.
Thus, exhaustive search turns out to be impractical
when modelling real-world situations using argumentative frameworks.
According to the definition of \emph{justification},
a dialectical tree resembles an {\sc and-or} tree: even
though an \Argum{\mathcal{A}}{h} may have many possible
defeaters \Argum{\mathcal{B}_1}{h_1}, \Argum{\mathcal{B}_2}{h_2}, \ldots,
\Argum{\mathcal{B}_k}{h_k},
it suffices to find just {\em one} acceptable
defeater \Argum{\mathcal{B}_i}{h_i}\ in order
to consider \Argum{\mathcal{A}}{h}\ as
defeated. Therefore, when analyzing the acceptance of
a given argument \Argum{\mathcal{A}}{h}\ not every node in
the dialectical tree \Tree{\Argum{\mathcal{A}}{h}}\ has to be expanded
in order to determine the label of the root.
\alfabeta\ pruning 
can be applied to speed up the labeling procedure,
as shown in figure~\ref{FigPodaAlfaBeta}(a).
Non-expanded nodes
are marked with an asterisk $\star$. Note: dialectical trees
are assumed to be computed depth-first.












\setlength{\unitlength}{0.6mm}

\begin{small}
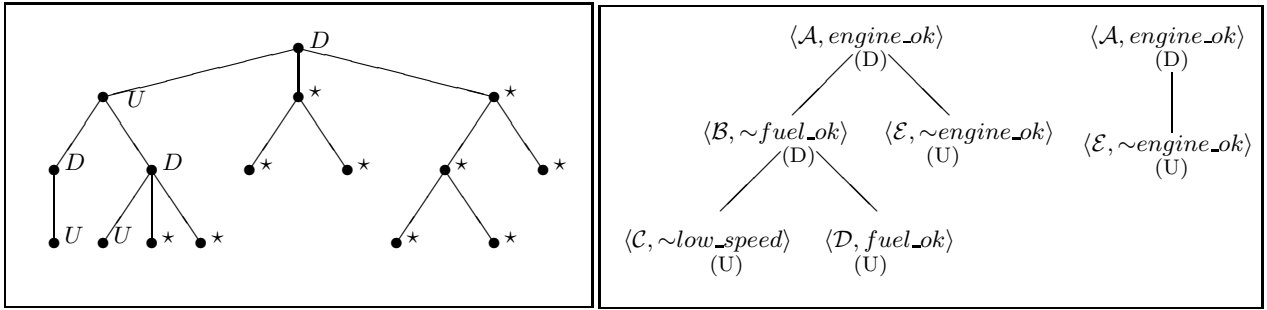
\begin{figure}
\begin{picture}(220,60)
\put(0,0){
\begin{picture}(120,60)
\setlength{\unitlength}{0.65mm}

\put(0, -3){\framebox(120,62){}}

\put( 60,50){\circle*{2} $D$}
\put( 60,50){\line(-4,-1){40}}
\put( 60,50){\line(4,-1){40}}
\put( 60,50){\line(0,-1){10}}
\put( 20,40){\circle*{2}}
\put( 25,38){$U$}

\put( 60,40){\circle*{2} $\star$}
\put(100,40){\circle*{2} $\star$}

\put( 20,40){\line(-2,-3){10}}
\put( 20,40){\line( 2,-3){10}}
\put( 10,25){\circle*{2} $D$}
\put( 30,25){\circle*{2} $D$}

\put( 60,40){\line(-2,-3){10}}
\put( 60,40){\line( 2,-3){10}}
\put( 50,25){\circle*{2} $\star$}
\put( 70,25){\circle*{2} $\star$}

\put(100,40){\line(-2,-3){10}}
\put(100,40){\line( 2,-3){10}}
\put(90,25){\circle*{2} $\star$}
\put(110,25){\circle*{2} $\star$}

\put(10,25){\line(0,-1){15}}
\put(10,10){\circle*{2} $U$}

\put(30,25){\line(0,-1){15}}
\put(30,25){\line(-2,-3){10}}
\put(30,25){\line( 2,-3){10}}
\put(20,10){\circle*{2} $U$}
\put(30,10){\circle*{2} $\star$}
\put(40,10){\circle*{2} $\star$}

\put(90,25){\line(-2,-3){10}}
\put(90,25){\line( 2,-3){10}}
\put(80, 10){\circle*{2} $\star$}
\put(100,10){\circle*{2} $\star$}

\end{picture}
}

\put(130,-10){
    \setlength{\unitlength}{0.8mm}
    \begin{picture}(110,55)
    \put(0,5){\framebox(110,50){}}
    \put(45,50){\makebox(0,0){\Argum{\mathcal{A}}{engine\_ok}}}
    \put(45,46){\makebox(0,0){{\scriptsize (D)}}}
    \put(42,47){\line(-1,-1){10}}
    \put(48,47){\line(1,-1){10}}
    \put(33,34){\makebox(0,0){\Argum{\mathcal{B}}{{\no}fuel\_ok}\hspace*{5mm}}}
    \put(33,30){\makebox(0,0){{\scriptsize (D)}}}
    \put(57,34){\makebox(0,0){\hspace*{8mm}\Argum{\mathcal{E}}{{\no}engine\_ok}}}
    \put(57,30){\makebox(0,0){{\scriptsize (U)}}}
    \put(30,31){\line(-1,-1){10}}
    \put(36,31){\line(1,-1){10}}
    \put(21,16){\makebox(0,0){\Argum{\mathcal{C}}{{\no}low\_speed}\hspace*{4mm}}}
    \put(21,12){\makebox(0,0){{\scriptsize (U)}}}
    \put(45,16){\makebox(0,0){\hspace*{6mm}\Argum{\mathcal{D}}{fuel\_ok}}}
    \put(45,12){\makebox(0,0){{\scriptsize (U)}}}

    \put(95,50){\makebox(0,0){\Argum{\mathcal{A}}{engine\_ok}}}
    \put(95,46){\makebox(0,0){{\scriptsize (D)}}}
    \put(95,44){\line(0,-1){10}}
    \put(95,32){\makebox(0,0){\Argum{\mathcal{E}}{{\no}engine\_ok}}}
    \put(95,28){\makebox(0,0){{\scriptsize (U)}}}

    \end{picture}
 }
\end{picture}
   \caption{(a) Labeling a dialectical tree with $\alpha-\beta$ pruning. \ \ (b) Dialectical trees (example~\protect\ref{EjemEngineTwo})}
    \label{FigPodaAlfaBeta}
    \label{FigDialecticalTrees}
\end{figure}
\end{small}

It is well-known that whenever \alfabeta\ pruning can be applied,
the {\em ordering} according to which nodes are expanded
affects the size of the search space 
Consider our former example: when determining whether
\Argum{\mathcal{A}}{engine\_ok}\ was justified, we computed depth-first
{\em all} arguments involved, thus obtaining the dialectical
tree \Tree{\Argum{\mathcal{A}}{engine\_ok}}\
 shown in figure~\ref{FigDialecticalTrees}(b)-left.
However,
had we started by considering the defeater \Argum{\mathcal{E}}{\neg engine\_ok}\
before than \Argum{\mathcal{B}}{\neg fuel\_ok},
we would have come to the same outcome by
just taking a subtree of \Tree{\Argum{\mathcal{A}}{engine\_ok}}
(as shown in figure~\ref{FigDialecticalTrees} (b)-right).
Computing this set exhaustively is a complex task,
since we should consider {\em every} possible counterargument
for \Argum{\mathcal{A}}{h}, determining whether
it is an acceptable defeater or not.
In order to formalize the {\em ordering} for expanding
defeaters as the dialectical tree is being built, we will
introduce a partial order $\preceq_{eval}$ as follows:

\begin{definition}
\label{DOrdenDeEvaluacion}
Let $S$ be a set of defeaters for \Argum{\mathcal{A}}{h}.
Given two arguments \Argum{\mathcal{A}_1}{h_1}\ and \Argum{\mathcal{A}_2}{h_2}\
in $S$, we will
say that $\Argum{\mathcal{A}_1}{h_1}\ \preceq_{eval} \Argum{\mathcal{A}_2}{h_2}$
iff \Argum{\mathcal{A}_1}{h_1}'s label
is computed before than \Argum{\mathcal{A}_2}{h_2}'s label.
\end{definition}

\begin{example}
In example~\ref{EjemEngineTwo}, it is the case
that $\Argum{\mathcal{B}}{{\no}fuel\_ok} \preceq_{eval}
\Argum{\mathcal{E}}{{\no}engine\_ok}$.
\end{example}

In dialectical trees, only {\em acceptable}
defeaters are considered, i.e. those which are non-fallacious
(as mentioned in example~\ref{EjemEngineTwo}).
Let \Argum{\mathcal{A}}{h}\ be an argument in a dialectical tree.
Then we will denote as $AcceptableDefeaters(\Argum{\mathcal{A}}{h})$
the set \{\Argum{\mathcal{B}_1}{h_1}, \ldots, \Argum{\mathcal{B}_n}{h_n} \}
of acceptable defeaters for \Argum{\mathcal{A}}{h}\ in that tree.

\begin{example}
Consider example~\ref{EjemEngineTwo}. It holds that
\Argum{\mathcal{B}}{{\no}fuel\_ok}\ is an acceptable
defeater for \Argum{\mathcal{A}}{engine\_ok}, whereas
\Argum{\{ \drule{{\no}oil\_ok}{heat} \} }{{\no}oil\_ok}
is {\em not} an acceptable defeater for
\Argum{\mathcal{E}}{{\no}engine\_ok}.
\end{example}

The algorithm in figure~\ref{FigConstruirArbolMejorado}
shows how a dialectical tree can be built and labelled in a depth-first
fashion, using both \alfabeta\ pruning and
the evaluation ordering $\preceq_{eval}$.
In order to speed up the construction of a dialectical tree,
our approach will be twofold. First, given an argument \Argum{\mathcal{A}}{h},
we will establish a syntactic criterion
for determining the set $AcceptableDefeaters(\Argum{\mathcal{A}}{h})$.
Second, we will give a definition of $\preceq_{eval}$ which prunes
the dialectical tree
according to consistency constraints.
Both approaches will be discussed in
section~\ref{SecCommitmentSet} and~\ref{SecSharedBasis}, respectively.

\subsection{Commitment set}
\label{SecCommitmentSet}


We will consider three distinguished sets of  literals associated
with an argument \Argum{\mathcal{A}}{h}: \\
(a) the set of {\em points for counterargumentation}
(literals which are conclusions of counterarguments for \Argum{\mathcal{A}}{h});\\
(b) the set of {\em points for defeat}
(literals which are conclusions of defeaters for \Argum{\mathcal{A}}{h}); and \\
(c) the set of {\em points for attack}
(literals which are conclusions of acceptable defeaters
for \Argum{\mathcal{A}}{h}\ in a given dialectical tree).

We will denote these sets as
$PointsForCounterarg(\Argum{\mathcal{A}}{h})$,
$PointsForDefeat(\Argum{\mathcal{A}}{h})$, and
$PointsForAttack(\Argum{\mathcal{A}}{h},\linarg)$, respectively.
From definitions~\ref{DefCounterargument} and~\ref{DefDefeater},
each of these sets is a subset of the preceding ones, i.e.:
$PointsForAttack(\Argum{\mathcal{A}}{h},\linarg) \subseteq PointsForDefeat(\Argum{\mathcal{A}}{h})
\subseteq PointsForCounterarg(\Argum{\mathcal{A}}{h})$

The set $PointsForAttack(\Argum{\mathcal{A}}{h},\linarg)$ represents the {\em optimal}
set of literals to take into account for building defeaters
for \Argum{\mathcal{A}}{h}, in the
sense that every literal in this set accounts for a conclusion
of an acceptable defeater.
In~\cite{Simari92}, the approach to determine
all possible defeaters for a given argument \Argum{\mathcal{A}}{h}\
considered the deductive closure of the
complement of the literals which
are consequents of those rules (in $\Pi$\ and $\cal A$) used in deriving $h$.
This notion, which will prove useful for pruning
the search space, will be characterized as {\em commitment set}:



\begin{definition}
\label{DCommit}
Let $\cal P$ = $(\Pi,\Delta)$ be a \dlp, and
let \Argum{\mathcal{A}}{h}\ be an argument in $\cal P$.
The {\em commitment set} of \Argum{\mathcal{A}}{h}\ wrt $\cal P$,
denoted $Commit({\Argum{\mathcal{A}}{h}})$, is defined as
$Commit({\Argum{\mathcal{A}}{h}}) =  \{ a \mid a\ {\rm is\ a\ ground\ literal\ such\ that\ \Pi\ \cup\  \Co{\mathcal{A}}} \vdash a \}$,
where
$Co(\cal A)$ denotes the set of consequents of defeasible
rules in $\cal A$.
If $S=\{\Argum{\mathcal{A}_1}{h_1}, \ldots \Argum{\mathcal{A}_n}{h_n}\}$
is a set of arguments,
then
$Commit(S)$ = \{ $a$ $\mid$ $a$ is a ground literal
such that $\Pi\ \cup\  \bigcup_{i=1\ldots n} \Co{\mathcal{A}_i} \vdash a \}$.
\end{definition}

The set $\overline{Commit(\Argum{\mathcal{A}}{h})}$\footnote{$\overline{S}$ stands for the set formed by the complement of every literal in $S$. E.g: $\overline{\{a,{\no}b\}} = \{{\no}a, b\}$.}
is suggested
in~\cite{Simari92} as an approximation
to $PointsForAttack(\Argum{\mathcal{A}}{h},\linarg)$.
From the preceding inclusion relationship, it follows that
$PointsForAttack(\Argum{\mathcal{A}}{h},\linarg) \subseteq\ \overline{Commit(\Argum{\mathcal{A}}{h})}$.
One of our goals is to find a better upper bound
for $PointsForAttack(\Argum{\mathcal{A}}{h},\linarg)$.
Next we introduce a lemma to
consider a proper subset of $\overline{Commit(\Argum{\mathcal{A}}{h})}$
for finding acceptable defeaters by backward chaining,
thus reducing the number
of defeaters to take into account.
That subset is given by the
\emph{consequents of defeasible rules} in $\mathcal{A}$.

%

\begin{lemma}\footnote{The lemmas in this paper are based on
the ones presented in~\cite{Chesne96} and~\cite{GarciaMSC97}.}
\label{PropFundamental}
Let \Argum{\mathcal{A}}{h}\ be an argument. Let \Argum{\mathcal{B}}{j}\ be an acceptable
defeater for \Argum{\mathcal{A}}{h}, i.e., \Argum{\mathcal{B}}{j}\ defeats \Argum{\mathcal{A}}{h}.
Then $B$ is also an argument for a ground literal $q$,
such that $q$ is the complement
of some consequent of a defeasible rule in $A$, and
\Argum{\mathcal{B}}{q}\ is an acceptable defeater.
\end{lemma}

Hence, we can find a better upper bound for the
set $PointsForAttack(\Argum{\mathcal{A}}{h},\linarg)$ by considering
the set $\overline{Co(A)}$.
Note that this set can be immediately computed once
the argument \Argum{\mathcal{A}}{h}\ has been built, whereas
the approach given in~\cite{Simari92} involved
computing the much more complex deductive
closure $(\Pi \cup\ \mathcal{A})^{\Infer}$.

\newcommand{\tab}[1]{\hspace*{#1mm}}

\setlength{\unitlength}{0.8mm}
\begin{figure}[thn]
\begin{small}
\begin{center}
\begin{picture}(120,80)
\put(0,0){\framebox(130,80){}}
\put(5,75){
\parbox[t]{120mm}{
\begin{Algo}{BuildDialecticalTree}  \ \\
\label{AlgoArbolDialectico}
{\sc Input:} \Argum{\mathcal{A}}{h} \\
{\sc Output:} \Tree{\Argum{\mathcal{A}}{h}}
{\em \{uses \alfabeta\ pruning  and
     evaluation ordering $\preceq_{eval}$ \}} \\
Let $S = AcceptableDefeaters(\Argum{\mathcal{A}}{h})$ \\
{\bf If} $S \ne \emptyset$ \\
\tab{3}  {\bf then} \\
\tab{7}    {\bf While} there is no $\Argum{\mathcal{A}_i}{h_i} \in S$ labelled as {\em U} \\
\tab{9}    {\bf For every} argument in $S$ \\
\tab{11}         Let $\Argum{\mathcal{A}_i}{h_i}$= minimal non-labelled element in $(S,\preceq_{eval})$ \\
\tab{11}            BuildDialecticalTree(\Argum{\mathcal{A}_i}{h_i}) getting as a result \Tree{\Argum{\mathcal{A}_i}{h_i}}\\
\tab{11}         Put \Tree{\Argum{\mathcal{A}_i}{h_i}} as a immediate subtree of \Argum{\mathcal{A}}{h}. \\
\tab{7}    {\bf If} there exists some \Tree{\Argum{\mathcal{A}_i}{h_i}} labelled as {\em U} \\
\tab{9}           {\bf then} Label \Tree{\Argum{\mathcal{A}}{h}}\ as {\em D} \\
\tab{9}           {\bf else} Label  \Tree{\Argum{\mathcal{A}}{h}}\ as {\em U} \\
\tab{3}  {\bf else} \\
\tab{7}     \Tree{\Argum{\mathcal{A}}{h}} = \Argum{\mathcal{A}}{h}, and
              Label  \Tree{\Argum{\mathcal{A}}{h}}\ as {\em U} \\
\end{Algo}
}}
\end{picture}
\end{center}
\end{small}
\caption{Algorithm for building and labelling a dialectical tree}
\label{FigConstruirArbolMejorado}
\end{figure}

\subsection{Commitment and evaluation order. Shared basis}
\label{SecSharedBasis}

As remarked in
sec.~\ref{Sec:DELP}, fallacious argumentation is to be avoided.
In \DELP, this means that all odd-level
(even-level) arguments in an argumentation line
\linarg = [\Argum{\mathcal{A}_0}{h_0},
           \Argum{\mathcal{A}_1}{h_1}, \ldots,
           \Argum{\mathcal{A}_k}{h_k}] must be {\em non-contradictory} wrt $\Pi$
to avoid {\em contradictory}  argumentation.
Def.~\ref{DCommit} captures the notion of commitment
set for an argument \Argum{\mathcal{A}}{h}. We will use that notion
for pruning the search space to determine
possible  defeaters for \Argum{\mathcal{A}}{h}, without considering
the whole set $\overline{\Co{A}}$.

 \begin{lemma}
 \label{LemaCommit}
 Let $\linarg$ be an argumentation line
 in  a dialectical tree $\Tree{\Argum{\mathcal{A}}{h}}$, such that
 $S_{\lambda}^{k}$ denotes the set of
 all supporting arguments  in $\lambda$ with level $\leq k$.
 Let $a$ be a ground literal, $a \in Commit(S_{\lambda}^{k})$.
  Let $\Argum{\mathcal{B}}{j} \in I_{\linarg}$, such that its level is greater
  than $k$. Then
  $\cpl{a} \not\in PointsForAttack(\Argum{\mathcal{B}}{j},\lambda)$.
 \end{lemma}

This lemma establishes the following:
assume that an argumentation line has been built up to level $k$.
If  an interferring argument were then introduced
at level $k' > k$, it could not be further attacked by a
supporting argument with conclusion ${\no}a$ at level $k'' > k'$,
if it is the case that $a$ belongs to
$Commit(S_{\lambda}^{k})$.
Thus, the former lemma accounts for the need of not falling into
`self-contradiction'  when an argument exchange is performed.
In order to introduce new supporting (interferring) arguments,
the proponent (opponent) is committed to what he has stated before.
This allows us to further reduce the set
of literals $\overline{Co(\Argum{\mathcal{A}}{h})}$ to take into account
for determining defeaters for \Argum{\mathcal{A}}{h}.
As an argumentation line is being built, if $a$
is a literal  in a supporting (interferring) argument
at level $k$, its complement $\cpl{a}$
cannot be the conclusion of supporting (interferring)
arguments at level $k' > k$.
As a direct consequence from lemma~\ref{LemaCommit},
literals present in {\em both} supporting
and interferring arguments up to level $k$ in a given argumentation line
{\em cannot be further argued at level $k'>k$}.

\begin{definition}[SharedBasis]
\label{DSharedBasis}
Let $\linarg = [ \Argum{\mathcal{A}_0}{h_0}, \ldots, \Argum{\mathcal{A}_n}{h_n} ]$ be an
argumentation line in \Tree{\Argum{\mathcal{A}}{h}}.
We define  $SharedBasis(\linarg,k)$
as the set of ground literals in the deductive
closure of:
 a) $\Pi$; b)
      the consequents of rules in both $S_{\lambda}$ and
       $I_{\lambda}$
       up to level $k$
       within the argumentation line $\linarg$.
 Formally:\footnote{If $S$ is a set
of arguments \{ \Argum{\mathcal{A}_1}{h_1},
\Argum{\mathcal{A}_2}{h_2}, \ldots,
\Argum{\mathcal{A}_k}{h_k} \},
then $DRules(S)$ denotes the set of all defeasible rules in $S$,
i.e., $DRules(S)$ = $A_1 \cup A_2 \cup \ldots \cup A_k$.}

\noindent
$SharedBasis(\lambda, k)   =  \{ a : a\ {\rm is\ a\ ground\ literal,\ and\ }
                              a \in (\Pi\ \cup (Co(DRules(S_{\lambda}^{k})) \cap
 (Co(DRules(I_{\lambda}^{k})))^{\vdash} \}$
 \end{definition}

From this definition we can state the following lemma,
which excludes literals belonging to the shared basis
(up to a given level $k$) as points for attack for arguments
at deeper levels.

\begin{lemma}[Commitment Lemma]
\label{LemaSharedBasis}
Let $a \in SharedBasis(\LAMBDA,k)$, $k \geq 0$.
Then  $\cpl{a} \not\in Points\-For\-Attack(\Argum{\mathcal{B}}{j},\LAMBDA)$,
for any argument $\Argum{\mathcal{B}}{j}\ \in \LAMBDA$.
\end{lemma}

From lemma~\ref{PropFundamental} and~\ref{LemaSharedBasis}
it follows that those literals belonging to $Co(A)$ which
are in $SharedBasis(\LAMBDA,k)$ {\em cannot be} the conclusions
of defeaters for \Argum{\mathcal{A}}{h}.
This allows us to get an improved upper bound for the
potential points for attack when computing
defeaters for a given argument \Argum{\mathcal{A}}{h}\ at level $k$ in
a dialectical tree.

 \begin{center}
 $PointsForAttack(\Argum{\mathcal{A}}{h},\linarg)
 \subseteq  \overline{Co(\Argum{\mathcal{A}}{h}) - SharedBasis(\LAMBDA,k)} \subseteq
 \overline{Co(\Argum{\mathcal{A}}{h})} \subseteq
\overline{Commit(\Argum{\mathcal{A}}{h})}$
 \end{center}

\subsection{Preference criterion}
\label{SecCriterio}

From the preceding analysis we can come back to the original
question: how to choose those defeaters belonging to the
most `promising' argumentation line? (i.e., those which
are more prone to break the debate as soon as possible).
From our preceding results, we can introduce the following definition
for $\preceq_{eval}$:

\begin{definition}[Evaluation ordering based on shared basis]
Let $\lambda$ = [ ...., \Argum{\mathcal{A}}{h}] be an argumentation line whose last
element is an argument \Argum{\mathcal{A}}{h}\ at level $k-1$.
Let \Argum{\mathcal{A}_{1}}{h_{1}}\ y \Argum{\mathcal{A}_{2}}{h_{2}}\ be
two possible defeaters for \Argum{\mathcal{A}}{h},
so that choosing
\Argum{\mathcal{A}_{1}}{h_{1}}\ would result in
an argumentation line
$\lambda_1$ = [ ...., \Argum{\mathcal{A}}{h},\Argum{\mathcal{A}_{1}}{h_{1}} ],
and choosing
\Argum{\mathcal{A}_{2}}{h_{2}}\ would result in
an argumentation line
$\lambda_2$ = [ ...., \Argum{\mathcal{A}}{h},\Argum{\mathcal{A}_{2}}{h_{2}} ].
Then
$\Argum{\mathcal{A}_{1}}{h_{1}} \preceq_{eval} \Argum{\mathcal{A}_{2}}{h_{2}}$\ iff \\
\begin{center}
$\overline{Co(\Argum{\mathcal{A}_{1}}{h_{1}}) - SharedBasis(\lambda_1,k)}$
$\subseteq$
$\overline{Co(\Argum{\mathcal{A}_{2}}{h_{2}}) - SharedBasis(\lambda_2,k)}$
\end{center}
\end{definition}

This evaluation order can be now applied in the
algorithm~\ref{AlgoArbolDialectico}.
An advantageous feature of this evaluation order
is that it is easy to implement. Given two
alternative defeaters for an argument \Argum{\mathcal{A}}{h},
the one which shares {\em as
many ground literals as possible} with the argument
(\Argum{\mathcal{A}}{h}) being attacked should be preferred,
thus maximizing the set $SharedBasis$.

\begin{example}
Consider examples~\ref{EjemEngineOne} and~\ref{EjemEngineTwo}.
Figure~\ref{FigDialecticalTrees} showed two alternative
ways of determining whether \Argum{\mathcal{A}}{engine\_ok}\
is a justification.
The consequents of defeasible rules are, in this case,
$Co(\mathcal{A})$ = \{ $engine\_ok$,
             $fuel\_ok$,
             $oil\_ok$,
             $pump\_fuel\_ok$,
             $pump\_oil\_ok$
          \}.
The argument \Argum{\mathcal{A}}{engine\_ok}\ has two
acceptable defeaters:
\Argum{\mathcal{B}}{{\no}fuel\_ok} and
\Argum{\mathcal{E}}{{\no}engine\_ok}.
In the first case,
$Co(\mathcal{B})$ = \{ $pump\_clogged$,
             $pump\_fuel\_ok$,
             $low\_speed$
          \},
and in the second case,
$Co(\mathcal{E})$ = \{ ${\no}engine\_ok$,
             $fuel\_ok$,
             $oil\_ok$,
             $pump\_fuel\_ok$,
             $pump\_oil\_ok$
          \}.
If we choose the defeater \Argum{\mathcal{B}}{{\no}fuel\_ok},
we have $\overline{Co(A)-SharedBasis(\lambda_1,1)}$
= \{ $\neg engine\_ok$, $\neg fuel\_ok$, $\neg oil\_ok$,
     $\neg pump\_oil\_ok$ \}.
Choosing the defeater
\Argum{\mathcal{E}}{{\no}engine\_ok}, we have
$\overline{Co(A)-SharedBasis(\lambda_2,1)}$
= \{ $\neg engine\_ok$ \}.
Since
$\overline{Co(A)-SharedBasis(\lambda_2,1)}$
$\subset$
$\overline{Cpl(Co(A)-SharedBasis(\lambda_1,1)}$,
the defeater
\Argum{\mathcal{E}}{{\no}engine\_ok}\ should be tried
before than \Argum{\mathcal{B}}{{\no}fuel\_ok}\
when computing the dialectical tree
\Tree{\Argum{\mathcal{A}}{engine\_ok}}.
\end{example}

\section{Conclusions and related work}
\label{sec:conclusions}

Defeasible Argumentation is a relatively new field in Artificial
Intelligence. Inference in argument-based systems is hard to tackle,
since its computational complexity is similar to related approaches, such
default logic~\cite{PraVre99}. Following the basic idea presented in this paper,
some experiments have been performed with examples of greater size,
although further research needs to be done in this area.

We contend that the approach presented in this paper gives a relevant
contribution to currently existing work~\cite{PraVre99,Vreeswijk97}.
Given two arguments \Argum{\mathcal{A}_1}{q_1}\
and \Argum{\mathcal{A}_2}{q_2}, other alternative formalizations
(such as Prakken and Sartor's~\cite{PraVre99} or Vreeswijk's~\cite{Vreeswijk97})
consider a full
consistency check $\Pi \cup\ \mathcal{A}_1 \cup\ \mathcal{A}_2 \vdash p, {\no} p$
to determine whether two arguments attack each other.
In this paper, we characterized attack in a goal-oriented way,
which rendered easier many implementation issues, and helped
to prune dialectical trees. It should be noted that \DELP~\cite{GarciaMSC97}
has been implemented using this goal-oriented attack, which resembles
the approach used by~\cite{KakasToni99} for normal logic programs.
However, \DELP\ provides richer knowledge representation capabilities,
since it incorporates both default and strict negation.\footnote{A full
analysis of these features is beyond the scope of this paper.}

Studying the need of avoiding {\em fallacious}
argumentation~\cite{Chile94}, we arrived at the notion of {\em commitment}
and {\em shared basis}, which allowed us to define a preference
criterion for dynamically obtaining the (on the average)
shortest argumentation lines when the justification procedure
is carried out.
Although our analysis was particularly focused on \DELP,
the approach presented in this paper can be adapted to many
existing argumentation  systems.

\bibliographystyle{alpha}

\end{document}